\documentclass[final,1p,times]{elsarticle}

\usepackage{lipsum}
\makeatletter
\def\ps@pprintTitle{%
	\let\@oddhead\@empty
	\let\@evenhead\@empty
	\def\@oddfoot{}%
	\let\@evenfoot\@oddfoot}
\makeatother

\usepackage{graphicx}
\graphicspath{ {images/} }
\usepackage{float}
\usepackage[utf8]{inputenc}
\usepackage{lmodern}
\usepackage{textcomp}
\usepackage{amsmath}
\usepackage{mathtools}
\usepackage[overload]{empheq}

\usepackage{siunitx}
\usepackage{color}

\usepackage{wrapfig}
\usepackage{graphicx}

\usepackage{picins}

\usepackage[pagewise]{lineno}

\begin{document}

\begin{frontmatter}

\title{Multi-Period Flexibility Forecast for Low Voltage Prosumers}

\author[label1,label2]{Rui Pinto}
 \author[label1]{Ricardo J. Bessa\corref{cor2}}
 \ead{ricardo.j.bessa@inesctec.pt}
\author[label1,label2]{Manuel A. Matos}
 \address[label1]{INESC Technology and Science (INESC TEC), Campus da FEUP, Rua Dr. Roberto Frias, 4200-465 Porto Portugal}
 \address[label2]{Faculty of Engineering, University of Porto, Rua Dr. Roberto Frias, 4200-465 Porto Portugal}
\cortext[cor2]{Corresponding author} 

\begin{abstract}

Near-future electric distribution grids operation will have to rely on  demand-side flexibility, both by implementation of demand response strategies and by taking advantage of the intelligent management of increasingly common small-scale energy storage. The Home energy management system (HEMS), installed at low voltage residential clients, will play a crucial role on the flexibility provision to both system operators and market players like aggregators. Modeling and forecasting multi-period flexibility from residential prosumers, such as battery storage and electric water heater, while complying with internal constraints (comfort levels, data privacy) and uncertainty is a complex task. This papers describes a computational method that is capable of efficiently learn and define the feasibility flexibility space from controllable resources connected to a HEMS. An Evolutionary Particle Swarm Optimization (EPSO) algorithm is adopted and reshaped to derive a set of feasible temporal trajectories for the residential net-load, considering storage, flexible appliances, and predefined costumer preferences, as well as load and photovoltaic (PV) forecast uncertainty. A support vector data description (SVDD) algorithm is used to build models capable of classifying feasible and non-feasible HEMS operating trajectories upon request from an optimization/control algorithm operated by a DSO or market player.

\end{abstract}


\end{frontmatter}

\section{Introduction}
\subsection{Motivation}

Distributed Renewable Energy Sources (DRES) have been experiencing a fast growing in medium voltage (MV) and low voltage (LV) grids as the solar power technology becomes more and more affordable \cite{ref_1_Hasheminamin2013, Taylor2016}. Conventional electrical network infrastructures were designed to accommodate unidirectional power flows coming from the large power plants to the more populating zones where most of the consumers exist. With the increasingly presence of DRES in LV and MV distribution grid, there is a paradigm change as the power flows start to reverse direction, particularly during sunny days or windy periods. Consequently, technical difficulties regarding the operation of distribution grid start to arise for Distribution Systems Operators (DSO), with bus voltage limits being violated or even line congestion events.

Microgrids, composed with flexible loads, small-scale storage and their intelligent management by means of an Home Energy Management System (HEMS) combined with smart meter capabilities, can bring flexibility into the operation of distribution grids, taking advantage of increasingly more frequent types of flexible loads such as the electric vehicle \cite{Alipour20171168, Anand2016241} and the presence of significant levels of photovoltaic (PV) microgeneration \cite{Clastres201055}. The flexible character that microgrids bring to the distribution grid operation can be used to provide ancillary services \cite{Biegel2014479, Majzoobi2017555} at period of grid stressful operation. Optimizing the operation of these small-scale distribution grids is being seen as crucial to endow DSOs of means of accommodate current levels of DRES and allow for a deeper penetration of micro-generation in LV distribution grids in the short-term future. Controlling, modeling, optimizing schedules of microgrid flexible assets has been topic of research for the last years \cite{LiangModelingMicrogrid,KimSchedulingMicrogrids} with some focus given to energy storage \cite{MilanesControlStorage,en9070528}.

DSOs can also take advantage of the flexible nature of microgrids during grid stressful operation periods, use it for voltage control features at MV/LV substations \cite{Tuominen2014}, for load frequency control \cite{Zhou_Load_Frequency_Control}, for power losses reduction, support unintentional microgrid operation \cite{Gouveia2017}, and to create operational conditions that maximize DRES hosting capacity that might bring financial benefits for prosumers and have a positive impact on the decarbonization of the electric power system.

Domestic small-scale storage, heat-pumps, thermostatically controlled loads (TCL), air-conditioners, and the electric vehicle will for sure be more common in residential type buildings in the near future, increasing the flexibility that can be provided within LV distribution grids, either by means of demand response programmes or flexibility aggregators participating in dedicated market sessions or even with single HEMS smart operation aiming at maximizing customers profits \cite{Shakeri2017154,Rajalingam2016184} while supporting DSO in meeting specific operational criteria \cite{ref_4_Snape2013, ref_5_Gkatzikis2013}. The flexibility provision from HEMS monitoring and control capabilities is a theme of relevant significance as a result of the added value that can be brought to interested agents like DSOs and flexibility aggregators. Therefore, there is a need for developing flexibility models to capture the following characteristics: 
\begin{itemize}
\item Multi-period flexibility from behind-the-meter storage and TCL due to the inter-temporal nature of state of charge (SoC) and water temperature equations;
\item LV net-load patterns driven by weather conditions (e.g., PV generation) that introduce high uncertainty in the forecasting task \cite{Qiu2017}. 
\end{itemize}

This paper produces contributions that address these two research challenges.

\subsection{Related Work and Contributions}

Different authors studied the economic and technical benefits of flexibility from energy storage technologies. The focus was mainly in large-scale storage and renewable energy power plants, which, combined, create the virtual power plant (VPP) concept. For instance, in \cite{ZAFIRAKIS2016971} the value of price arbitrage for energy storage is evaluated for different European electricity markets. The authors found that in fuel import-based markets, such as the United Kingdom, there are opportunities for investors to benefit from energy storage arbitrage and the presence of hydro power plants decrease the interest in storage investment. Gomes et al. proposed a two-stages stochastic optimization method for the joint coordination of wind and PV systems considering energy storage in the day-ahead market and adopted a mixed-integer linear programming formulation \cite{GOMES2017310}. Results show that, although the total amount of traded energy is inferior when considering a joint coordination, the final profit is greater than when considering a disjoint operation. A comprehensive literature review about VPP scheduling can be found in \cite{Nosratabadi2017}.

The present paper differs from these works since it is focused in residential prosumers with small-scale storage and PV generation, and flexible appliances combined through an HEMS. In this segment, concerns like data protection and privacy, multi-period flexibility representation and the need to have local computational units are important challenges that are not fully covered by the current literature.

A standard approach for modeling flexible distributed energy resources (DER) is to characterize their flexibility with a set of specific parameters. For instance, for planning the operation of services buildings with thermal energy storage on a yearly basis, Stinner et al. defined the maximum flexibility provision in each hour considering three dimensions: temporal, power and energy \citep{STINNER2016140}. However, in this work the flexibility was modeled and calculated for planning purposes rather than a short-term time horizon. For design process or for selecting set of buildings to participate in demand response schemes, De Coninck et al. proposed flexibility cost curves, which correspond to the amount of energy that can be shifted to or from a flexibility interval and the associated cost \cite{DECONINCK2016653}. Operational decision making is not covered by this methodology.

Multi-period flexibility modeling and forecasting for short-term horizons is a rather recent topic, with few relevant research works dedicated to it. 

Zhao \textit{et al.} studied a geometric approach that is capable of aggregating flexibility provided from TCL (represented as a ``virtual'' battery), and where the set of feasible power profiles of each TCL has been demonstrated to be a polytope \cite{ref_8_Zhao2016}. The aggregation of several sets is performed by means of the Minkowski sum of the individual polytopes. The computational burden issue is tackled by the authors by adopting several approximations regarding the calculation of the Minkowski sums. Despite the merits of this work, domestic battery storage is not modeled and neither is the impact of net-load uncertainty, which are both relevant and challenging issues in this problem.

The ``virtual'' battery model was also explored by other authors, like Hughes \textit{et al.} that proposed a first-order linear dynamical model for flexibility provision from heating, ventilation, and air conditioning (HVAC) systems in frequency regulation services, and generalizes the method to many other types of loads \cite{Hughes2016}. The results showed that the developed technique still has challenges to overcome in modeling small-scale systems. Another example is Hao \textit{et al.}, which considered demand aggregation using battery models to model the set of feasible power profiles that a collection of TCLs can provide to track frequency regulation signals~\cite{Hao2015}. The TCL modeling assumes a simplified continuous power model where the error related to this simplification decreases as the size and homogeneity of the TCL aggregation increases.

Nosair \textit{et al.} proposed a method to construct flexibility envelopes that describe the flexibility potential of the power system and its individual resources \cite{ref_12_Nosair2015}. The proposed envelopes comprise all possible intra-hourly deviation and variation of the modeled DRES considering that for a certain sub-hourly time there is maximum output variability. Using the 95\% percentile of the probability distribution of all the sub-hourly time steps the authors define an envelope which comprises the majority of realizations of flexibility requirements for that intervals. HEMS are not considered in this study, particularly the costumer's preferences regarding the operation of their equipment, which makes the modeling problem more complex and simultaneously more realistic. Moreover, the multi-period nature of flexibility is not modeled in the envelope. 

A similar concept is also proposed by Nuytten \textit{et al.} \cite{ref_11_Nuytten2013}. The authors presented a methodology to estimate the maximum and minimum curves regarding the operation of a combined heat and power (CHP) plant combined with thermal storage. The difference between these two curves is indicated by the authors as being the theoretical maximum flexibility that the system can provide. This methodology can only be used for modeling the maximum flexibility that the system is capable of providing for one specific time step in the time horizon considered, if one assumes that no flexibility has already been provided in a later period. No multi-temporal formulation has been adopted in this work, which means that a set of feasible power set-points regarding flexibility provision during more than one time step cannot be provided by the proposed method. This flexibility representation was also adopted in the European Project IndustRE as \textit{flexgraphs} \cite{Delnooz2016}.

Ulbig and Anderson \cite{Ulbig2015a} described a methodology to analyze the available flexibility for each time-step from an ensemble of diverse units in a confined grid zone. This flexibility is modeled as a Power Node, which allows for detailed modeling of specific constraints such as maximum ramp rates, power limits, as well as energy storage operation ranges. The authors propose a visual representation of the available flexibility during a specific time horizon. Nevertheless, this visual representation and modeling approach does not account for a multi-temporal formulation, meaning that the mentioned flexibility availability is only depicted for a single time step.

Pan et al. proposed a method to calculate the feasible region of a linear programming formulation for the operation of a district heating system, considering the thermal inertia of buildings \cite{PAN2017395}. The main objective was to have condensed and privacy preserving data exchange between district heating and electricity control centers. The method can be generalized to estimate the flexibility region for multiple time periods and non-linear formulations, but showed the following limitations: a) uncertainty is not included in this work (is pointed out by the authors as future work); b) visual representation of  multi-period flexibility region with more than three time intervals is not possible. 

A different representation for the flexibility is proposed in  \cite{Bremer2013}. Bremer and Sonnenschein propose two sampling methods for defining the technically feasible flexibility set from DER. The authors, for instance, propose a Monte Carlo sampling method that starts with a feasible operating schedule and then, in each step, modifies it in at least one point in time with a random mutation factor. The methodology is used by the authors in a succeeding work where the obtained trajectories are used as a learning sample for a support vector data description (SVDD) algorithm \cite{Sonnenschein2015}. Despite some similarities in the methodology with the work being presented in the present paper, this approach did not consider the modeling of costumers' preferences neither accounted for the forecast uncertainty.

Considering the reviewed state of-the-art, the main original contribution from the present paper is a novel trajectory generation algorithm that is capable of modeling the multi-period flexibility from HEMS, including information about base net-load (i.e., inflexible load plus PV generation) forecast uncertainty represented by a set of short-term scenarios generated from probabilistic forecasts taking into account the temporal interdependency of forecast errors.

The main limitation of reviewed literature is that information about net-load forecast uncertainty is not included in the multi-period flexibility model, which, when included in DSO or aggregators optimization models, might lead to solutions with low robustness to uncertainty. In contrast to the methods described in \cite{ref_8_Zhao2016, Hughes2016, Hao2015, PAN2017395, Bremer2013} that proposed multi-period flexibility estimation but without forecast uncertainty characterization, in our method the uncertainty in the PV and net-load profile is tackled by means of a novel trajectory-based evaluation procedure using the convergence features of the EPSO (Evolutionary Particle Swarm Optimization \cite{EPSO_Miranda}) algorithm that have been adapted to this work.

Compared to \cite{STINNER2016140, DECONINCK2016653}, the proposed method is for operational (or short-term decision-making) and extends flexibility estimation to multiple time intervals (i.e., set of flexibility trajectories), which represents an added value compared to the works described in \cite{ref_12_Nosair2015, ref_11_Nuytten2013, Delnooz2016, Ulbig2015a}.

Similarly to \cite{PAN2017395, Bremer2013}, the modeling strategy allows the interested parts to not have to model the equipment within the HEMS, which reduces the computational complexity and effort of problems such as multi-period OPF and maintains data privacy. In a second stage, the flexibility trajectory set is used as input in a SVDD model that is capable of delimiting the feasible flexibility set of the respective HEMS. The potential interested parts, DSOs or demand/flexibility aggregators, only need to receive a reduced number of flexibility trajectories, called support vectors, which by means of a specific function to be embedded in their optimization tools allows them to identify unknown HEMS flexibility trajectories as being feasible or not.

\subsection{Structure of the Paper}

The remaining of this paper is organized as follows: section \ref{Concept} introduces the concept of multi-temporal flexibility adopted in this work; in section \ref{Meth} the developed methodology is detailed, presenting the structure of the flexibility set generation algorithm, and the approach to encode and distribute the generated information to interested parts; section \ref{res} presents the results regarding the performance of the proposed method; finally, in section \ref{conc} the main conclusions are presented.

\section{HEMS Multi-period Flexibility: the Concept}
\label{Concept}

Multi-period flexibility from HEMS can be defined as the ability to change the expected (baseline) net-load profile for a specific period of time (e.g., 24 hours period), by jointly considering information about flexible/inflexible load, PV generation, hot water demand, water temperature inside the electric water heater and the state of charge (SoC) of domestic battery storage. 

The visual representation of the multi-period flexibility envelope is not as straightforward as one might assume. Actually, when dealing with problems that aim at defining the feasible flexibility set for more than three time steps ahead, the visual representation of such domain becomes impossible. There is a difference between the visual representation of the power limits in each time step that characterize the maximum flexibility band (like the flexibility envelope in \cite{ref_12_Nosair2015, ref_11_Nuytten2013}) and the actual visual representation of the feasible flexibility set considering different temporal activations of flexible resources.

\begin{figure}[H]
	\centering
	\includegraphics[width=10cm]{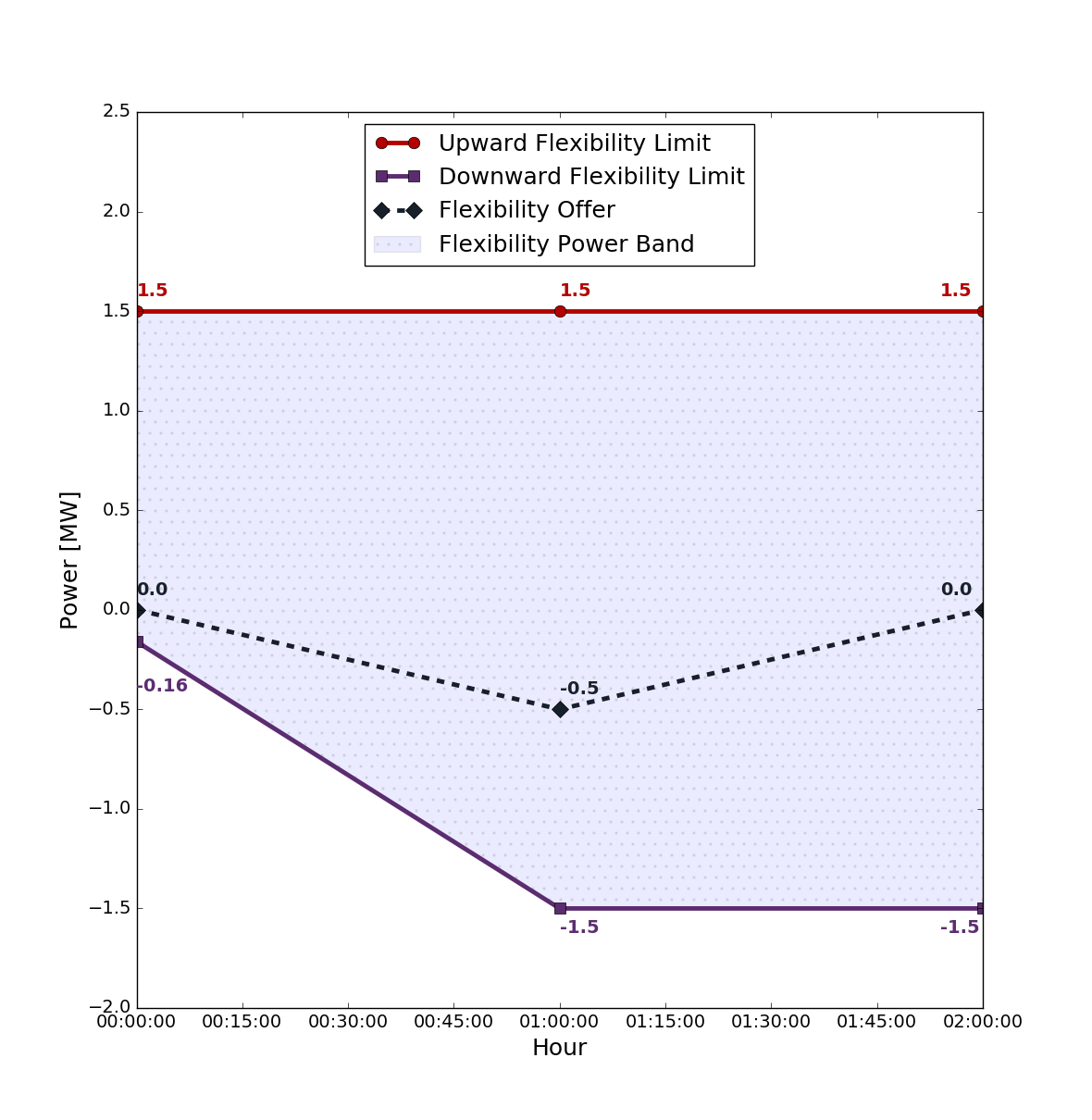}
	\caption{Hourly limits for flexibility search space.}
	\label{fig:search_space}
\end{figure}

For the sake of clarity, let one assume that in the following example the flexibility provision can only be provided by a single domestic battery storage with 3.2 kWh of electrical energy capacity, maximum charge and discharge power of 1.5 kW, an initial SoC of 0.64 kWh (20\%) and a minimum allowed SoC of 15\% capacity, 0.48 kWh. In this example the battery efficiency is neglected. The upward and downward hourly limits of the flexibility band for this battery, considering the stated conditions, are depicted in Figure \ref{fig:search_space}. This would represent the flexibility envelope or \textit{flexgraph} from \cite{ref_12_Nosair2015, ref_11_Nuytten2013}).  

The outer limits of the flexibility domain represented in Figure \ref{fig:search_space} are defined by the previously mentioned upward and downward flexibility limits. The search domain defined in Figure \ref{fig:search_space} relates to what is commonly referred to as the flexibility power band. As the initial SoC is 0.64 kWh, the upward flexibility limit for the first considered operating time is equal to the maximum battery charging power, as with that the resulting SoC will be 2.14 kWh, considering hourly time steps and neglecting the battery's efficiency. Analyzing the defined downward flexibility limits, namely the limit for the first time step considered, as the minimum allowed SoC is 15\% of the storage capacity, 0.48 kWh, the maximum discharge power admissible is just 0.16 kW for a hourly time step.
From the second operation time step considered onwards, both the upward and downward flexibility provision power limits correspond to the maximum charge and discharge rates, respectively. This occurs because there is always a possible flexibility trajectory that remains feasible while representing a choice of maximum charging or discharging power in any of the remaining time steps considered.

With that said, it is important to stress that, as previously stated, there is an important difference between the flexibility power band limits definition and the limits for the feasible flexibility provision envelope being tackled in this study. An example of that fact is the trajectory being depicted also in Figure \ref{fig:search_space} representing a possible flexibility offer expressed in kW [0.0, -0.5, 0.0]. Although the power set-points composing this trajectory are all within the limits of the defined flexibility power band, analyzing the SoC response to such trajectory (in kWh) [0.64, 0.14, 0.14] one can verify that the trajectory becomes infeasible from the second time step onwards as the minimum SoC constraint (SoC $>=$ 0.48 kWh) is not being complied.

The representation for the flexibility space that is adopted in this paper is through a set of technically feasible net-load trajectories, which represent alternative paths to the expected (baseline) net-load profile (trajectory). In other words, these trajectories are samples taken from the multi-dimensional space forming the feasible flexibility set.  

Concluding, the concept of multi-temporal flexibility provision relates with the potential that a certain HEMS has of reshaping its expected net-load profile for a determined number of inter-temporal related periods, while complying with technical and physical internal constraints. The main focus of this study refers to the delimiting of the flexibility set that encompasses all the possible multi-temporal net-load profile variations that can be performed by the HEMS control functions. The next section describes the methodology that generates this set of temporal flexibility trajectories.

\section{Methodology for Modeling the Feasible Flexibility Set}
\label{Meth}

\subsection{General Framework}

The methodology developed in this work extends the previous work reported in \cite{Pinto2017} in which the feasible flexibility set is estimated using semi-randomly generated feasible trajectories and then feeding a SVDD algorithm with those trajectories. In that previous version of the algorithm, random sampling routines were being used to generate a sufficient number of feasible trajectories. In this new version, the construction of feasible trajectories no longer depends on a random sampling routine but instead an EPSO algorithm is being used to search for feasible trajectories. The use of the EPSO algorithm also enables the inclusion of information about base net-load uncertainty forecast (i.e., inflexible load plus PV generation) by means of solution evaluation for a set of different uncertainty scenarios, which greatly increases the complexity and computational effort. Accordingly, a feasible solution will be one that complies with all the constraints for a predefined probability threshold from all the possible HEMS base net-load scenarios considered, instead of using simply a point (or deterministic) forecast information like in the previous algorithm version.

The final set of feasible trajectories resulting from the EPSO search procedure are aggregated to create a learning dataset for the SVDD algorithm. The SVDD is an one-class support vector machine algorithm that is commonly used in novelty detection, where a determined set of samples is provided to the function which in turn builds a model by detecting the soft boundary of that set \cite{ref_15_Pedregosa2012}. Inspired by the methodology proposed in in \cite{ref_16_Bremer2011} for the encoding of search spaces for virtual power plants application, the SVDD is used in this work for classifying new flexibility trajectories as belonging to that set or not. Or in other words, to check if a potential net-load profile is technically feasible or not.

\begin{figure}[H]
	\centering
	\includegraphics[width=6cm]{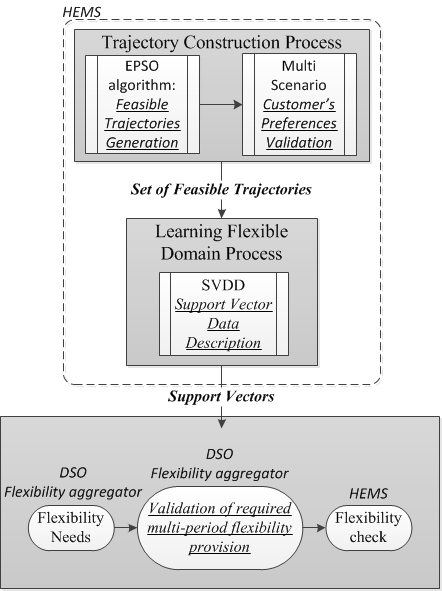}
	\caption{Block-diagram of the flexibility set search algorithm.}
	\label{fig:flowchart}
\end{figure}

Figure \ref{fig:flowchart} depicts the main stages of the proposed methodology in the form of a block-diagram. The first step is the generation of short-term scenarios for the forecast uncertainty, corresponding to a discrete multi-temporal representation of uncertainty, which is described in section \ref{sec:uncertainty}.
Then, there are three main boxes representing: the trajectory construction process; the learning of the feasible domain process; and the validation of the multi-period flexibility. The first stage concerns to the use of the EPSO algorithm to generate feasible trajectories that comply with the defined customer's preferences (see section \ref{sec:epso}). These customer's preferences are embedded into the proposed algorithm and are responsible of modeling the desire of minimizing the wasting of energy coming from PV generation, which means that the battery's storage capacity must be used at its most to accommodate the energy surplus from PV generation. Another considered customer preference is the definition of the water temperature range inside the EWH tank, which must never surpass the defined minimum and maximum temperatures during the period of time in study. The formulation of the optimization problem is discussed in section \ref{sec:formulation}. 

In the second stage, section \ref{black_box}, the constructed feasible trajectories are used as input for the SVDD function which will result in the model construction and identification of the support vectors that define the boundaries of the feasibility domain. Finally, in the validation of multi-period flexibility stage, the interested agent (DSO or flexibility aggregator) can take advantage of the feasible domain knowledge coming from the information embedded in the provided support vector and define the optimal multi-period flexibility trajectory that is aligned with its operational needs while being viable to be provided by the HEMS/aggregation of HEMS.

\subsection{Representation of Forecast Uncertainty}\label{sec:uncertainty}

%
%
The proposed method to generate flexibility trajectories for HEMS uses as input a sample of temporal scenarios without any assumption of the parametric model of its joint distribution. Each scenario can be interpreted as a sample taken from the joint probability distribution. This representation is called \textit{random vectors} in statistics \cite{Cario1997}, \textit{path forecasts} in econometrics \cite{Jorda2010} and \textit{weather ensembles} in atmospheric sciences \cite{Bessa2017}.

Mathematically, a set of $M$ temporal scenarios for a time horizon of length $T$ can be defined as follows:
\begin{equation} \label{eq:trajectory}
{\bf Y}^M  = \left[ {\begin{array}{*{20}c}
   {y_{t + 1|t}^1 } & {y_{t + 2|t}^1 } &  \cdots  & {y_{t + T|t}^1 }  \\
   {y_{t + 1|t}^2 } & {y_{t + 2|t}^2 } &  \cdots  & {y_{t + T|t}^2 }  \\
    \cdots  &  \cdots  &  \cdots  &  \cdots   \\
   {y_{t + 1|t}^m } & {y_{t + 2|t}^m } &  \cdots  & {y_{t + T|t}^m }  \\
\end{array}} \right]
\end{equation}
where each row ($\bf y^{\left[ m \right]} $) of ${\bf Y}^M$ contains one scenario member. Collectively, the scenario set should exhibit the correct temporal dependence-structure structure between the marginal probability distributions or probabilistic forecasts (see \cite{Pinson2009} for more details).

These scenarios can be empirical, analytical, physical or generated in a stochastic model. Yet, it is important to underline that the proposed method is independent from the scenario generating process and can be applied even if this process is unknown or known only in the form of a simulation model.

Empirical scenarios can be constructed with analog-based methods like the analog ensemble approach \cite{eckel.2016}, which searches the historical forecast data for situations when the forecast was most similar (or analogous) to the current forecast. For each of those analogous forecasts, the corresponding observation is collected. An analytical solution to construct these scenarios is the epi-spline basis functions, which approximates the stochastic process for renewable energy and controls the degree to which extreme errors are captured \cite{Staid2017}. A well-know physical approach consists in ensemble predictions systems that are designed to model three sources of weather uncertainty: initial conditions, physical approximation and boundary conditions \cite{Bessa2017}.

In this paper, a stochastic simulation method, based on the Gaussian copula proposed in \cite{Pinson2009, Golestaneh2016} for wind and solar energy, was adopted. The method work as follows: 
\begin{itemize} 
\item Probabilistic forecasts (marginal distribution functions) for solar power are generated with a combination of feature engineering and gradient boosting trees and using a grid of weather forecasts \cite{Andrade2017}. Exponential functions are used for the distribution's tails as described in \cite{Matos2016}. For load time series, probabilistic forecasts are generated with conditional kernel density estimation \cite{Reis2017}.
\item Scenarios (or random vectors) are generated by plug-in an exponential covariance matrix into a Gaussian copula and using the inverse of the forecasted cumulative distribution function. The details of the scenario generation method can be found in Appendix A.
\end{itemize}

\subsection{Problem Formulation}\label{sec:formulation}

In this work, flexibility from domestic EWH and battery storage is included in the flexibility model. However, the methodology can be easily generalized to other flexible resources at the domestic and network level. The problem formulation has two decision variables and two state variables. The decision variables are the power flow in the domestic electric battery's inverter, \(P_{bat}\), and the operating point of the EWH, \(P_{ewh}\). The two state variables refer to the battery SoC and the water temperature inside the EWH tank.
The constraints used in this problem formulation are presented next.

{\begin{equation} \label{eq:1}
    traj\textsubscript{h} = Pbat\textsubscript{h} + Pewh\textsubscript{h}
\end{equation}

\begin{equation} \label{eq:2}
    Pbat^{min} <= Pbat\textsubscript{h} <= Pbat^{max}
\end{equation}

\begin{equation} \label{eq:3}
    Pewh\textsubscript{h}=
    \begin{cases}
      0, &\text{for off status}\ \\
      Pewh^{nom}, &\text{for on status}\
    \end{cases}
\end{equation}

\begin{equation} \label{eq:4}
    SoC^{ini} + \sum_{h=1}^{H}Pbat\textsubscript{h} <= SoC^{max}
\end{equation}

\begin{equation} \label{eq:5}
    SoC^{ini} + \sum_{h=1}^{H}Pbat\textsubscript{h} >= SoC^{min}
\end{equation}

\begin{equation} \label{eq:6}
    \theta^{min} <= \theta\textsubscript{h} <= \theta^{max}
\end{equation}

The trajectories representing the flexibility that can be provided by the HEMS are limited to the battery's charging and discharging powers (\(P\textsubscript{bat}^{max}\) and \(P\textsubscript{bat}^{min}\)) and the EWH nominal power (\(P\textsubscript{ewh}^{nom}\)), (\ref{eq:2}) and (\ref{eq:3}) respectively. 

Regarding the maximum charging power, in this study a dynamic model is adopted where the maximum charging power depends on the SoC of the battery, which is a typical behavior for Li-ion batteries. This is explained by the two most common charging stages that occur when charging a lithium-ion battery, constant current and constant voltage \cite{page_batteries}. During constant current charging stage the battery is basically connected to a current-limited power supply until reaching around 70-80\% of its energy capacity. For superior SoC the battery enters the constant voltage stage where the charger acts as a voltage limited power supply and the charging current gradually decreases as the SoC approximates full capacity. 

The modeling used in this work is presented in Figure \ref{fig:battery} where the maximum charging power starts to decrease for SoC superior to 80\% and a minimum charging power of 20\% nominal power is assumed. The inclusion of this non-linear behavior of battery storage also highlights the added value of the proposed method since it is not constrained by first-order linear models like in \cite{Hughes2016}.

\begin{figure}[H]
	\centering
	\includegraphics[width=6cm]{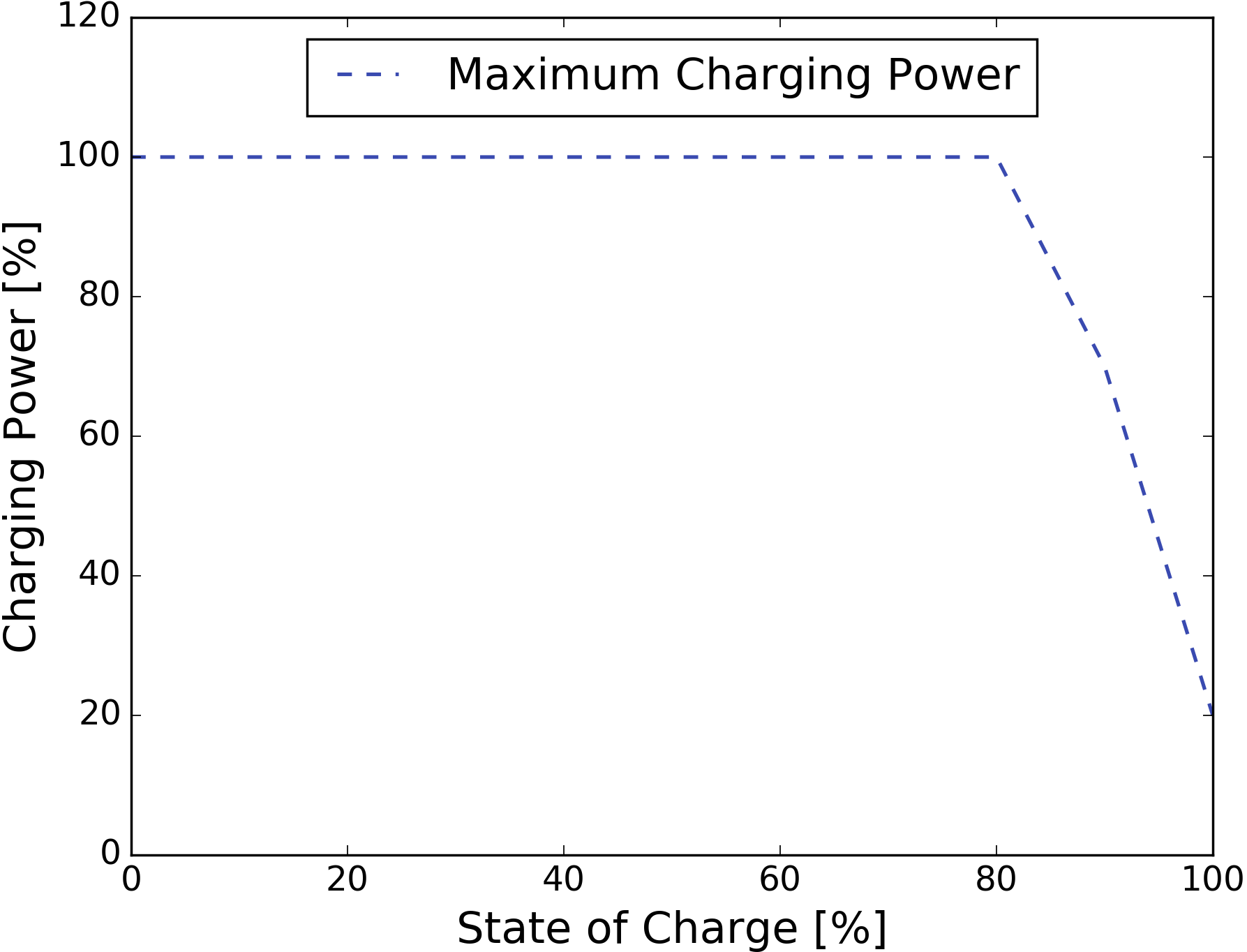}
	\caption{Battery Charging Model}
	\label{fig:battery}
\end{figure}

State of charge limits are enforced by equation (\ref{eq:4}) and (\ref{eq:5}), with the maximum allowed SoC being the total energy capacity of the battery and the minimum allowed SoC being limited to a certain percentage of total capacity, e.g. 15\%. There is an allowed water temperature range inside the EWH tank that is being represented by Equation (\ref{eq:6}). Equation (\ref{eq:7}) represents the water temperature variation along the time horizon, which depends on the expected volume of hot water usage and the decision variable regarding the operating status of the EWH, \(P\textsubscript{ewh}\). The physically-based load model adopted for the EWH modeling is aligned with the one used in \cite{Heleno2015}.

\begin{equation} \label{eq:7}
    \theta\textsubscript{h} = \theta\textsubscript{h-1} + \frac{\Delta t}{C} \left [ -\alpha(\theta\textsubscript{h-1} - \theta\textsubscript{house}) - c\textsubscript{p} v\textsubscript{h} (\theta\textsubscript{des}-\theta\textsubscript{ini})+ Pewh\textsubscript{h} \right ]
\end{equation}

In (\ref{eq:7}), \(\Delta t\) is the time step [\textit{h}], C is the thermal capacity [kWh/\textdegree C] set to 0.117, \(\alpha\) is the thermal admittance [kWh/\textdegree C] set to \(-9.42^{-4}\), \(\theta\)\textsubscript{house} is the house indoor temperature set to 20 \textdegree C, c\textsubscript{p} is the water specific heat [kWh/(ltr.\textdegree C)], v\textsubscript{h} is the hot water consumption at time h, \(\theta\)\textsubscript{des} is the desired temperature for water consumption set to 38 \textdegree C, and \(\theta\)\textsubscript{inl} is the inlet water temperature set to 17 \textdegree C.

In line with Figure \ref{fig:flowchart}, there is a need to validate the costumers preferences. Thus, the SoC state variable must be updated taking into account the PV generation surplus, which must be accommodated by the battery, accounting for battery storage capacity and maximum charging power limitations. Accordingly, for each time step of the operation horizon considered, the SoC variable is updated by summing up the PV generation surplus, \(PV^{sur}\), which consequently increases the SoC, and subtracting the EWH possible power in those time steps. The combination between the PV generation surplus to be accommodated, the decision variable regarding the charging (or not) of the battery, and the decision variable representing the EWH operating status must respect the maximum charging power, by verifying (\ref{eq:8}).

\begin{equation} \label{eq:8}
    Pbat_{h} + PV_{h}^{sur} - Pewh_{h} <= Pbat^{max}, \text{  }\forall h
\end{equation}

Additionally, (\ref{eq:4}) must give place to (\ref{eq:9}) to account for the PV generation surplus. The maximum and physically possible amount of PV surplus energy that the battery can absorb without being used for flexibility provision must still be assured when defining the feasible trajectories for flexibility provision.

\begin{equation} \label{eq:9}
    SoC^{ini} + \sum_{h=1}^{H}Pbat\textsubscript{h} + PV\textsubscript{h}^{sur} - Pewh_{h} <= SoC^{max}
\end{equation}

There is a maximum amount of PV generation surplus that the battery is capable of accommodating, which is related to its storage capacity and it can vary along the time horizon depending on the precedent operation decisions. The maximum energy that the battery can accommodate results from the difference between the maximum and minimum SoC limits. To assess whether this specification is being respected or not along all the time steps considered, an auxiliary variable was created, \textit{capacity}. Its initial value is set to the previously referred maximum energy that the battery can accommodate. The created auxiliary variable tracks down the supposed PV surplus energy accommodation capacity for each time step.

\begin{equation} \label{eq:update_cap1}
    capacity_{h} = capacity_{h-1} - (PV\textsubscript{h}^{sur} - Pewh_{h})
\end{equation}
\begin{equation} \label{eq:update_cap2}
    capacity_{h} = capacity_{h-1} + Pbat^{max}
\end{equation}

In each time step where there is a PV generation surplus, its value is updated by subtracting its current value by the PV surplus to be accommodated in that time step (\ref{eq:update_cap1}), limited to the battery charging power. On the other hand, in each time step where there is no PV generation surplus, its value is updated by increasing the capacity to absorb by the discharging power limit of the battery (\ref{eq:update_cap2}). A trajectory is classified non-feasible if is limiting the theoretical maximum capacity of the storage unit to accommodate PV surplus. If a certain trajectory respects all the problem constraints besides the one regarding the accommodation of PV, that trajectory is modified to cope with that requirement.

\begin{equation} \label{eq:update_traj1}
    Pbat_{h} = capacity_{h} - (PV\textsubscript{h}^{sur} - Pewh_{h})
\end{equation}

Accordingly, the decision variable regarding the operation of the electric battery will be modified to be equal to the remaining battery capacity after accommodating the PV surplus (\ref{eq:update_traj1}). Part of the PV surplus might be consumed by the operation of the EWH while providing flexibility, which is accounted in (\ref{eq:update_cap1}) and (\ref{eq:update_traj1})

Figure \ref{fig:solar} illustrates the evolution of the state variable regarding the battery SoC after this evaluation procedure takes place.

\begin{figure}[H]
	\centering
	\includegraphics[width=10cm]{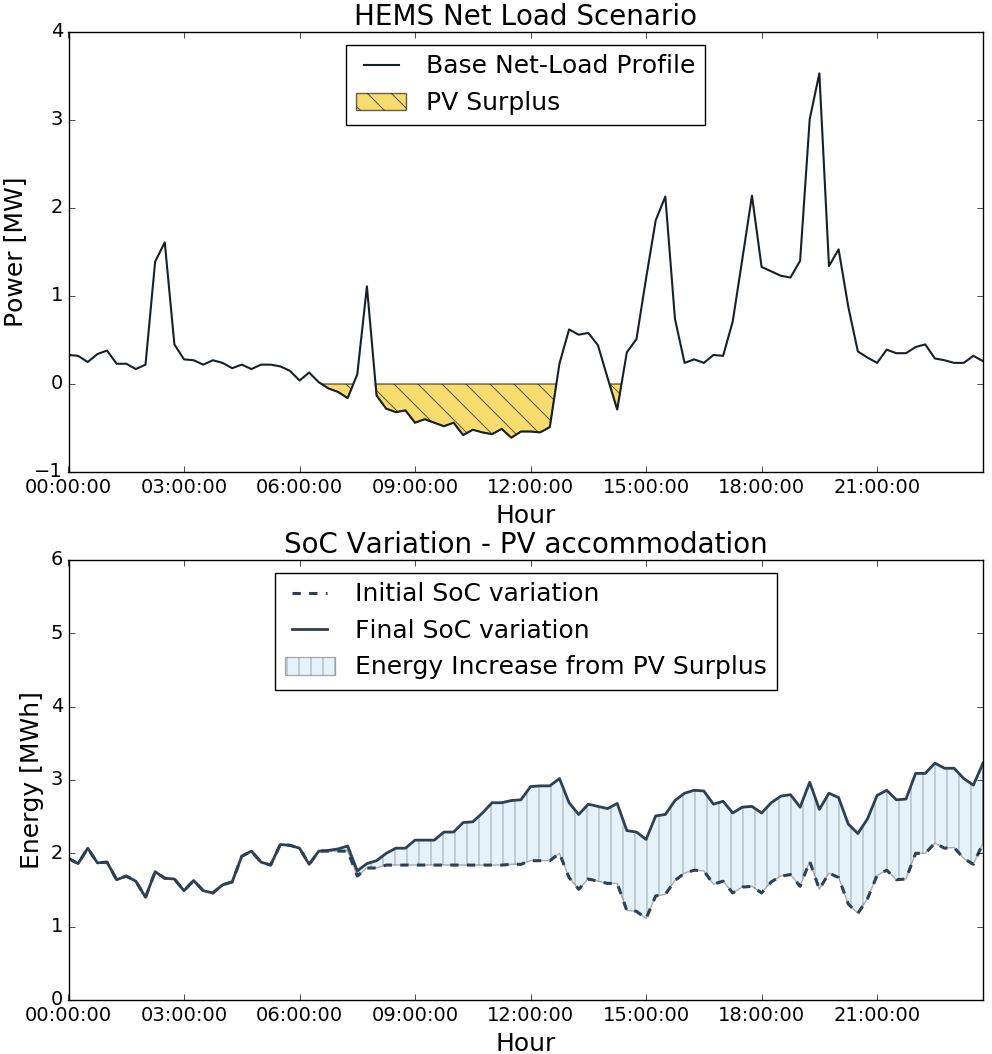}
	\caption{HEMS base net-load scenario (top); SoC variation due to PV surplus (bottom).}
	\label{fig:solar}
\end{figure}

Figure \ref{fig:solar} represents, at the top, one of the 100 possible base net-load scenarios. As previously mentioned, to meet the costumer's preferences, the battery SoC must be updated whenever the PV generation exceeds the load levels (represented by negative net-load values). One consequence of this costumer preference modeling is that, in moments of PV surplus, the battery cannot present downward flexibility (only possible with battery discharging) since the battery must be used to accommodate the energy PV surplus through the charging mode (although HEMS downward flexibility might be presented depending on the operation of the EWH). An example of this effect can be observed in Figure \ref{fig:solar}. The initial SoC variation depicted in the bottom chart shows that the SoC either stays unchanged or increases during the identified PV surplus moments. At the bottom, the resulting SoC variation along the considered time horizon can be compared to the original variation. As it can be seen from the figure analysis, the final SoC starts to differ from the initial SoC variation at the moment when the first net-load negative value occurs. The shape difference represents the PV generation surplus deducted the power consumed by the EWH in the flexibility provision.

\subsection{EPSO Implementation for the Trajectory Searching Process}\label{sec:epso}

The fundamental ideas behind the EPSO algorithm are the population based evolutionary programming concept where each combination of generated solution, \textit{X}, and respective strategic parameters, \textit{weights - w}, is called \textit{particle}. There are five main steps in the general scheme of EPSO, namely:

\begin{itemize}
\item \textbf{Replication:} where each particle is replicated;
\item \textbf{Mutation:} where each particle has its weights, \textit{w}, mutated;
\item \textbf{Reproduction:} where an offspring is generated from each mutated particle according to the movement rule;
\item \textbf{Evaluation:} where each particle in the population has its fitness evaluated;
\item \textbf{Selection:} where, by means of stochastic tournament, the best particles survive to form the next generation.
\end{itemize}

For a given particle \(X_{i}\), the new resulting particle, \(X_{i}^{new}\), results from:

\begin{equation} \label{eq:xi}
    X_{i}^{new} = X_{i} + V_{i}^{new}
\end{equation}

\begin{equation} \label{eq:vi}
    V_{i}^{new} = w_{i0}^{*}V_{i} +  w_{i1}^{*}(b_{i} - X_{i}) + w_{i2}^{*}(b_{g}^{*} - X_{i})
\end{equation}

This movement rule has the terms of inertia, memory and cooperation. The weights are subjected to mutation:

\begin{equation} \label{eq:wik}
    w_{ik}^{*} = w_{i1} + \tau N(0, 1)
\end{equation}

where N(0, 1) is a random variable with Gaussian distribution with 0 mean and variance 1. Additionally, the global best \(b_{g}\) comes randomly disturbed:

\begin{equation} \label{eq:bg}
    b_{g}^{*} = b_{g} + \tau^{'} N(0, 1)
\end{equation}

In (\ref{eq:wik}) and (\ref{eq:bg}) the \(\tau\) and \(\tau^{'}\) are learning parameters.

Using the EPSO method, the developed algorithm incorporates in each particle information regarding the decision variables,  \(P_{bat}\) and \(P_{ewh}\). Accordingly, each particle in this reshaped EPSO algorithm becomes a two-dimension object representing the two decision variables. The final fitness value of each particle, which represents the trajectory feasibility verification, is the result of the combined fitness evaluation of the two dimensions of each particle. During the fitness evaluation process the state variables are updated for the time steps considered. In the end, each feasible trajectory, \textit{traj}, will result from the sum, for each time step, of the two dimensions of each of the selected particles, according to (\ref{eq:1}). Problem constraint enumerated in the previous section must be complied. 

In each iteration of the algorithm, as part of the EPSO fitness evaluation process, the state variables must be assessed regarding the generated particles. Accordingly, the battery SoC and the water temperature of the EWH are computed based on the values of the previous time steps for the entire time horizon considered. As the developed EPSO algorithm considers a two-dimensional formulation, this state variable assessment procedure includes two independent functions. The trajectory feasibility relies on the fulfillment of (\ref{eq:5}), (\ref{eq:8}), and (\ref{eq:6}). For each time step where (\ref{eq:5}), (\ref{eq:8}), and (\ref{eq:6}) are not respected, a penalty term is added in the penalty verification function of the particle being evaluated (\ref{eq:fitness}).

\begin{equation} \label{eq:fitness}
    Penalty\textsubscript{k} = Pen\textsubscript{soc\textsubscript{max}} + Pen\textsubscript{soc\textsubscript{min}} + Pen\textsubscript{temp\textsubscript{water}}
\end{equation}

In line with the information presented in Figure \ref{fig:flowchart}, the feasible trajectories generation process that takes place in the developed EPSO algorithm is not complete without the customer's preferences validation. In this study, besides the assurance that the water temperature inside the EWH tank remains within the pre-established temperature range (\ref{eq:6}), one must assure that the main propose of the battery use prior to the HEMS flexibility offering remains being the accommodation of the PV generation surplus (\ref{eq:update_cap1}) and (\ref{eq:update_cap2}).

The validation of this customer requirement is accomplished by a scenario based approach that allows the EPSO resulting feasible flexibility set (trajectories) to incorporate the forecast uncertainty regarding the base net-load of the HEMS. Hence, a set of short-term scenarios are used to represent forecast uncertainty of base load and PV generation in this methodology. This probabilistic information is included in the constraints of the optimization problem, resulting in a chance-constrained optimization problem \cite{Charnes1959} that is solved with EPSO. Let $\varsigma$ be the indicator function on the fulfillment of constraint (\ref{eq:fitness}), as represented by (\ref{eq:indicator}). 

\begin{equation} \label{eq:indicator}
    \varsigma=
    \begin{cases}
      1, &\text{if Penalty\textsubscript{k} = 0}\ \\
      0, &\text{if Penalty\textsubscript{k} $>$ 0}\
    \end{cases}
\end{equation}

The indicator function $\varsigma$ can be seen as a binary indicator that gets value 1 if the penalty verification function in (\ref{eq:fitness}) equals 0 (i.e., no constraints violation), and has the value 0 if the the penalty verification function results in a value grater that 0 representing constraints violation. In other words, if the constraints are violated in one interval of the net-load scenario, $\varsigma$ gets value 0 independently from the violation or not of the constraints in other intervals, which is guaranteed by (\ref{eq:fitness}). The EPSO fitness function comes from the sum of the indicator function for all the considered base net-load scenarios (\ref{eq:fit_uncertainty}). Accordingly, and as defined in (\ref{eq:fit_uncertainty_final}), for a trajectory to be considered robustly feasible it must comply with a scenario percentage threshold $\tau$, e.g., the number of base net-load scenarios in which the trajectory remains feasible must be greater than a determined percentage of the total number of scenarios.

\begin{equation} \label{eq:fit_uncertainty}
    Fit\textsubscript{k}^{uncert} = \sum_{s=1}^{Nscen}\varsigma\textsubscript{s}
\end{equation}

\begin{equation} \label{eq:fit_uncertainty_final}
    Fit\textsubscript{k}^{uncert} > \tau*Nscen
\end{equation}

The trajectory fitness evaluation can be carried out according to (\ref{eq:fitness}) for the different conditions that each of the considered base net-load scenario represents. Consequently, one can classify a certain trajectory as being or not robustly feasible by checking (\ref{eq:fit_uncertainty}).

In each EPSO iteration, the ever encountered best global particle (i.e., the one with better fitness evaluation) needs to be updated. As referred, the fitness function used in this problem formulation only penalizes solutions that do not comply with the defined constraints. This means that there is no optimal solution in this searching process for which the EPSO algorithm is being used. Each particle that is feasible for all the considered scenarios will have the same fitness value. Therefore, the best global particle, \(b_{g}\), is selected based on the relative position of all the so far identified feasible trajectories. The aim here is that \(b_{g}\), which is used as reference in the movement rule, is chosen to increase the diversity of the feasible trajectories set. Accordingly, this selection procedure evaluates the relative position of all the elements in the set of feasible trajectories, looking for the one with the greatest distance relative to all the others. 

\begin{equation} \label{eq:glob_best}
	\centering
    Distance = \sum_{t=1}^{T}\left ( \left | P_{bat, t} - P_{bat, t}^{mean} \right | \right ) + \left ( \left | P_{ewh, t} - P_{ewh, t}^{mean} \right | \right )
\end{equation}

Thus, for each dimension and for all the time steps considered, the absolute distance between the particle position and the mean position of the feasible set is computed. The final distance comes from the accumulated distance along the time horizon considered, following (\ref{eq:glob_best}).

\subsection{Surrogate Model for the HEMS Flexibility}
\label{black_box}

The final EPSO algorithm output will be a large set of feasible temporal trajectories, which represent the feasible flexibility domain. The originated trajectories are to be used as a learning set of a SVDD function. The one-class support vector machine function available in the Scikit-Learn Python Library \citep{ref_15_Pedregosa2012} was used. This learning dataset must have sufficient diversity among the built trajectories so that the resulting model is capable of efficiently delimit and learn the feasibility domain boundary.

The trained SVDD model identifies the necessary support vectors (domain boundary representative trajectories) that describe the high-dimension sphere representing the feasible domain and the respective coefficients. The support vectors together with the respective coefficients compose the data that is transmitted to the interested agents in order to quantify the HEMS flexibility potential. From the entire set of feasible trajectories that feed the SVDD function, some are selected by it regarding the significance that they have on delimiting the feasible domain. During this identification process, support vectors coefficients are also computed, which imposes more or less significance on certain support vectors. This means that some support vectors are more decisive on the delimiting of the feasible domain, which implies that the respective coefficients have a greater value. Applying (\ref{eq:11}) the SVDD model is capable of classifying new flexibility trajectories as feasible or not. 

\begin{equation} \label{eq:11}
	R^{2}(x)=1-2\sum_{i}\beta\textsubscript{i}k(x\textsubscript{i}, x) + \sum_{i,j}\beta\textsubscript{i}\beta\textsubscript{j}k(x\textsubscript{i}, x\textsubscript{j})
\end{equation}

This classification is based on the comparison of the radius of the high dimension sphere and the radius in the high dimension domain that the trajectory being classified represents. The formula that calculates the correspondent trajectory (and sphere's) radius is expressed in equation (\ref{eq:11}), where $R^2$ is the square of the radius being calculated, $x_i$ and $x_j$ are support vectors, $\beta_i$ and $\beta_j$ are the respective coefficients, $k$ refers to the kernel type used by the SVDD function, and $x$ is the trajectory being evaluated. More detailed information regarding this methodology can be consulted in \cite{ref_16_Bremer2011} and \cite{Sonnenschein2015}. The information in equation (\ref{eq:11}) can  be integrated in any meta-heuristic optimization framework (like EPSO) to optimize the availability flexibility according to the end-user goals \cite{Bremer2014}.

To be classified as feasible, a trajectory must represent a radius in the high dimension domain that is equal or inferior to the radius of the sphere representing the feasibility boundary. Figure \ref{fig:svdd2} illustrates the SVDD classification procedure where the sphere representing the feasible flexibility space is defined by the identified support vectors. Trajectories whose projection falls within the flexibility space delimited by the sphere (in other others, trajectories that lead to radius smaller than the sphere's radius) are considered feasible. On the other hand, trajectories leading to radius greater that the sphere's radius are considered unfeasible, which in Figure \ref{fig:svdd2} are represented by the triangle-shaped symbols outside the domain defined by the sphere.

\begin{figure}[H]
	\centering
	\includegraphics[width=10cm]{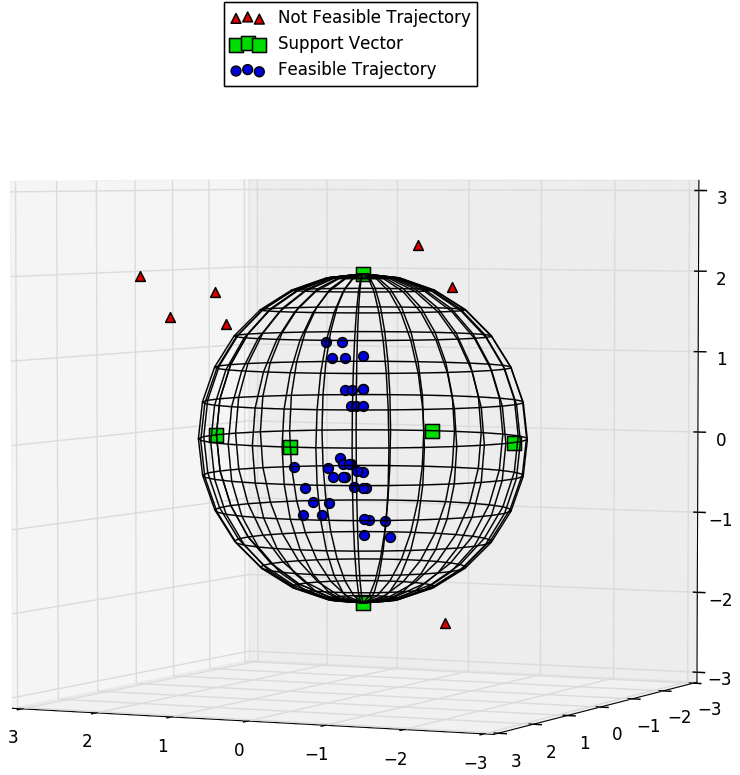}
	\caption{SVDD flexibility space concept}
	\label{fig:svdd2}
\end{figure}

The discrete nature of the EWH operation brings some challenges when defining the HEMS flexibility set, namely due to discontinuities that can be introduced by it. These discontinuities can be learnt by the SVDD algorithm, which might lead to misguiding trajectory classification. In order to clearly illustrate this effect, let one consider a EWH and a electric battery as flexible assets inside a HEMS. If at a certain time step the battery, for reasons of its operating strategy, does not have flexibility provision capability, the only flexibility that the HEMS can provide comes from the EWH discrete operation. Thus, the SVDD function will learn that for the referred time step the HEMS can provide upward flexibility by the amount equal to the nominal power of the EWH. The resulting trajectory classification model will then consider as feasible flexibility provision values between 0 kW and the EWH nominal power, instead of a discrete representation. Nevertheless, these operation conditions are not frequent and this modeling glitch can be neglected when the battery has enough SoC margin to adjust its power output.

According to the methodology introduced by this work, in order to define its flexibility potential the HEMS only needs to provide to the interested parts the computed support vectors and the respective coefficients, following (\ref{eq:11}), where $x_i$ and $x_j$ are the support vectors and $\beta_i$ and $\beta_j$ the respective coefficients. Accordingly, no information regarding customer's demand patterns or installed equipment (like battery specifications) need to be revealed. This surrogate model for the HEMS flexibility complies with recent concerns regarding costumer's data privacy arising with the smart meter deployment \cite{private_1, private_2, private_3}.

\section{Numerical Results}
\label{res}
The performance assessment of the developed methodology was based on two main analyzes: the generation of feasible trajectories and the classification accuracy of SVDD models.

\subsection{Generation of Feasible Flexibility Trajectories}
As detailed in section \ref{Meth}, a trajectory will be classified as feasible when complying with the considered constraints for at least a pre-defined minimum percentage of base net-load scenarios. The constraints refer to battery's SoC limits, EWH water temperature and the use of the battery during PV surplus periods. The EPSO algorithm is used to generate the set of feasible trajectories that will feed the SVDD function responsible to construct a model capable of identifying new trajectories as feasible or not.

For the base net-load forecast uncertainty of the HEMS, 100 short-term scenarios were used. In this study, for a certain trajectory to be considered feasible, it must comply with the problem constraints for at least 90 scenarios, which represents a probability of 90\%. The time horizon corresponds to a 24 hours window with a 15 minutes resolution, leading to 96 time steps.

The EPSO algorithm was configured with a 30 particle population size, a maximum number of iterations of 5000, a feasible trajectories target of 1000, communication factors of 0.15 for both dimensions of the particles, maximum and minimum mutation rates of 0.50 and 0.05, respectively, and a learning parameter, \(\tau\), of 5. The mutation rate is dynamic throughout the process, beginning with the maximum value and decreasing until the minimum value as the number of iterations increase. 

For helping in the convergence of the algorithm, an initial population is used, instead of using random values. This initial population fills the initial particles with decision values that respect the limits for the respective decision variables, \(P_{bat}\) and \(P_{ewh}\). Additionally, using one of the scenarios of HEMS net load profile, the PV energy surplus periods were identified and used to zero the decision variable \(P_{bat}\), so it approximates some of these initial particles to the costumer's preference constraint of not discharging the battery during those periods and use it to accommodate the referred PV energy surplus.

The HEMS modeled in this study has as flexible assets a domestic battery and an EWH unit. The battery considered has 3.2 kWh capacity, maximum charging and discharging power of 1.5 kW and an efficiency of 92.5\%. The initial SoC was set to 60\% of the maximum capacity and the minimum SoC level was defined as 15\% of the maximum capacity. Regarding the EWH, it has a nominal power of 0.5 kW, maximum and minimum water temperatures of 80\textdegree C and 45\textdegree C, respectively, and an initial water temperature of 60\textdegree C.

Figure \ref{fig:all_scens} depicts the 100 HEMS base net-load scenarios used in this study (top-left corner) and also a set of 100 feasible flexibility trajectories generated by the EPSO algorithm (bottom-left corner).

\begin{figure}[H]
	\centering
	\includegraphics[width=\textwidth,height=250pt]{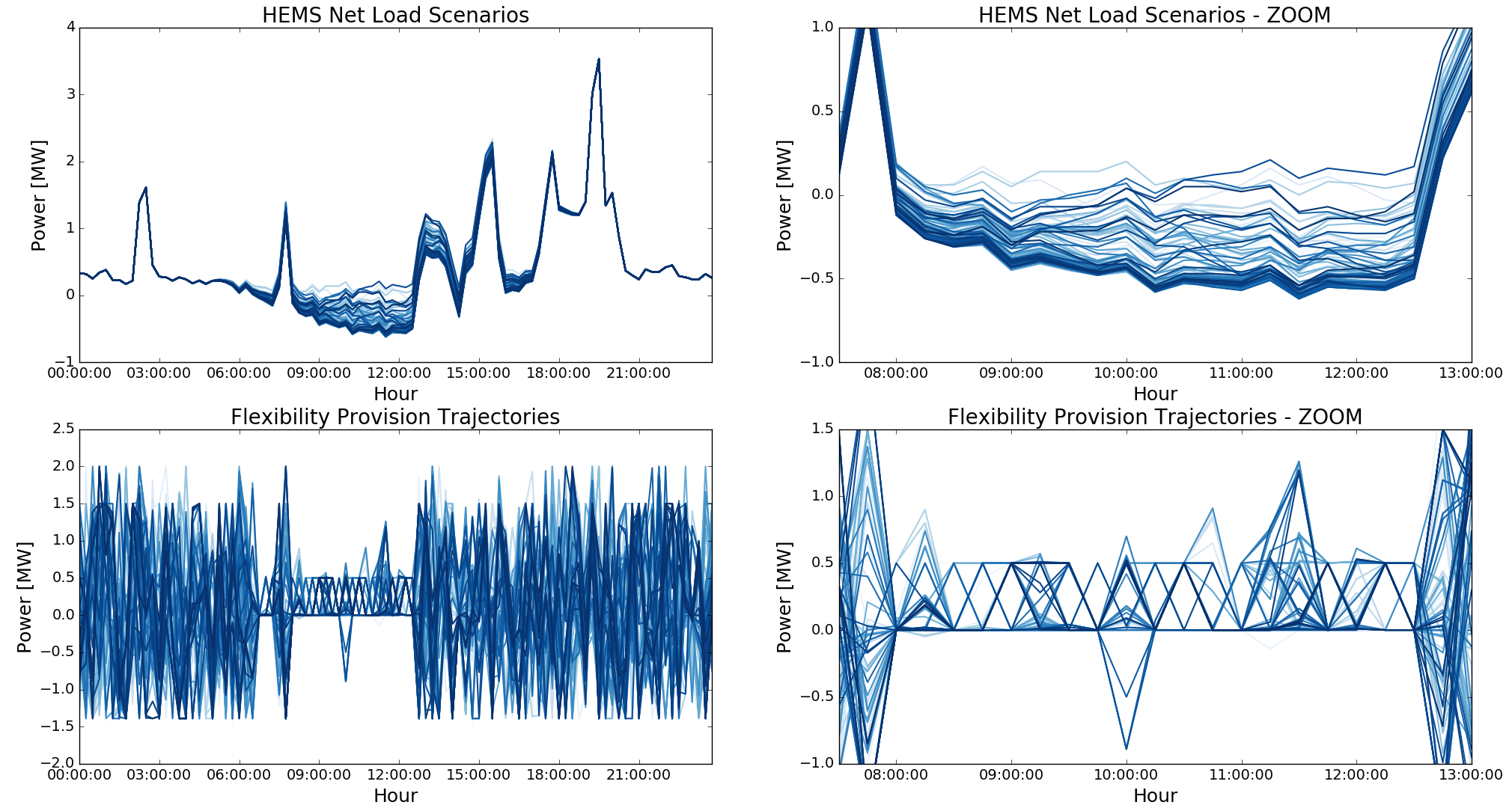}
	\caption{HEMS net-load trajectories (top); Flexibility provision trajectories (bottom).}
	\label{fig:all_scens}
\end{figure}

The net-load profile scenarios analysis (Figure \ref{fig:all_scens} top-left corner) lets one identify the period of day of typical superior PV generation as the one that brings more uncertainty regarding the HEMS net-load. It is during this period that the HEMS net-load can present negative values, which relate to the costumer's preference constraint of using the battery to minimize, as far as it is physical possible to the battery, the injection of PV energy in the grid. The impact of this constraint in the set of feasible trajectories constructed can be  observed in the bottom charts depicted in Figure \ref{fig:all_scens}. The period between 07:30 and 13:00 has been zoomed in for improved clarity (right hand side charts of Figure \ref{fig:all_scens}). 

From 08:00 to 12:30 most of the net-load scenarios represent a power injection in the LV distribution grid, which imposes that the battery should not be used to provide downward flexibility. The right hand side bottom chart shows the flexibility trajectories for that period. As one can observe, most of the trajectories are providing 0 kW or 0.5 kW, the latter referring to the EWH nominal power. Values greater that 0.5 kW can occur if the respective trajectory has room to provide upward flexibility, while accommodating the PV energy surplus. Negative values are very uncommon and only can occur if when evaluating a certain particle for the use of the battery costumer preference the EWH power related decision variable counterbalances the negative net-load trajectories in some time steps. If that occurs, the battery can be used freely during such time steps to provide flexibility.

Regarding the increase in the diversity of the solutions generated when comparing the current version of the algorithm with the preliminary one that used semi-random routines for the trajectory construction \cite{Pinto2017}, an analysis was performed using the principle component analysis method, which applies an orthogonal transformation to convert a set of variables (with a set of observations) into a set of values of linearly uncorrelated variables (principal components). Basically, the aim is to verify which version of the algorithm produces a 1000 set of feasible trajectories that needs more components to explain a certain percentage of the variance of the respective produced set. Results show that the newest algorithm version presented in this work needs 5 components to explain 50\% of the variance and 16 components to explain 80\%, while, for the same conditions, the older version needs only 2 and 5, respectively. This proves that the set of feasible trajectories produced by the newest version of the algorithm is more diverse, which leads to a better representation of the feasible search space when using the computed trajectories as input to build the SVDD model.

The final set of 1000 feasible trajectories generated by the EPSO routine needed around 25 minutes to be constructed in a desktop computer with an Intel Core i7-2600 CPU running at 3.40 GHz and with 8.00 GB of installed RAM. The algorithm was developed on Python programming language.

\subsection{Classification Accuracy of SVDD-based Model}

The other analysis carried out regarding the performance of the algorithm refers to the assessment of the SVDD classification accuracy of the multi-period trajectories. With that objective, two different sets of trajectories were used: one set composed by feasible trajectories and a second set of unfeasible trajectories. The purpose of using these two sets relates to the necessity of evaluating the classification process not only for the correct classification of feasible trajectories but also for correctly identifying unfeasible trajectories.

Regarding the SVDD function, hyper-parameters had to be defined before the construction of the classification models. These hyper-parameters are the kernel type and coefficient (\( \gamma \)), which will influence the quality of the classification performance. For this problem, it was found that the most suitable type of kernel was the \textit{Sigmoid}, being presented in Table \ref{tab:1} the performance results for that kernel type together with the results from models using the \textit{radial basis function} and the \textit{polynomial} kernels. Additionally, it was found that the fine tuning of the \textit{nu} parameter can have a strong influence on the quality of the constructed SVDD model, which can be consulted in Table \ref{tab:1}. For better model performance, the input trajectories were normalized for values between 0 and 1, regarding the minimum and maximum values from the EPSO feasible trajectories set. The time period used for this analysis ranged between 09:00 and 13:00 with 15 minutes time steps, resulting in a 16 time steps problem.

\begin{table}[H]
\centering
\caption{Model Efficiency Comparison for Feasible Trajectories Set}
\label{tab:1}
\begin{tabular}{||l|c|c|c||c|c|c||}
\hline
\multicolumn{7}{||c||}{\textbf{$\gamma$ = 0.05 and \textit{nu} = 0.01}}                                                                                                                                                      \\ \hline
\multicolumn{1}{||c|}{} & \multicolumn{3}{c||}{Feasible Trajectories Set}                                                         & \multicolumn{3}{c||}{Non-Feasible Trajectories Set}                                                          \\ \hline
\multicolumn{1}{||c|}{} & \# correct                & \# incorrect              & error (\%)                 & \# correct                & \# incorrect               & error (\%)                 \\ \hline
\textit{rbf}                    & 4964                      & 48                        & 0.96                       & 32                        & 11350                      & 99.72                      \\ \hline
\textit{poly}                   & 4960                      & 52                        & 1.04                       & 3929                      & 7453                       & 65.48                      \\ \hline
\textit{sigm}                & 4961                      & 51                        & 1.02                       & 3867                      & 7515                       & 66.03                      \\ \hline\hline
\multicolumn{7}{||c||}{\textbf{$\gamma$ = 0.05 and \textit{nu} = 0.1}}                                                                                                                                                                                                                                                                                                           \\ \hline
\multicolumn{1}{||c|}{} & \multicolumn{3}{c||}{Feasible Trajectories Set}                                                         & \multicolumn{3}{c||}{Non-Feasible Trajectories Set}                                                          \\ \hline
\multicolumn{1}{||c|}{} & \#correct                 & \# incorrect              & error (\%)                 & \#correct                 & \# incorrect               & error (\%)                 \\ \hline
\textit{rbf}                      & 4511                      & 501                       & 10.0                       & 496                       & 10886                   & 95.64                      \\ \hline
\textit{poly}                    & 4510                      & 502                       & 10.02                      & 9013                      & 2369                       & 20.81                      \\ \hline
\textit{sigm}                & 4510                      & 502                       & 10.02                      & 9027                      & 2355                       & 20.69                      \\ \hline\hline

\multicolumn{7}{||c||}{\textbf{$\gamma$ = 0.05 and \textit{nu} = 0.15}}                                                                                                                                                                                                                                                                                                          \\ \hline
\multicolumn{1}{||c|}{} & \multicolumn{3}{c||}{Feasible Trajectories Set}                                                         & \multicolumn{3}{c||}{Non-Feasible Trajectories Set}                                                          \\ \hline
\multicolumn{1}{||c|}{} & \#correct                 & \# incorrect              & error (\%)                 & \#correct                 & \# incorrect               & error (\%)                 \\ \hline
\textit{poly}                    & 4258                      & 754                       & 15.04
					    & 9632                      & 1750                      & 15.38 \\ \hline
\textit{sigm}                 & 4261                      & 751                       & 14.98
					    & 9624                      & 1758                      & 15.45 \\ \hline\hline

\multicolumn{7}{||c||}{\textbf{$\gamma$ = 0.05  and \textit{nu} = 0.20}}                                                                                                                                                                                                                                                                                                          \\ \hline
\multicolumn{1}{||c|}{} & \multicolumn{3}{c||}{Feasible Trajectories Set}                                                         & \multicolumn{3}{c||}{Non-Feasible Trajectories Set}                                                          \\ \hline
\multicolumn{1}{||c|}{} & \#correct                 & \# incorrect              & error (\%)                 & \#correct                 & \# incorrect               & error (\%)                 \\ \hline
\textit{poly}                    & 4010                      & 1002                      & 19.99
					    & 9928                      & 1454                      & 12.77 \\ \hline
\textit{sigm}                 & 4011                      & 1001                      & 19.97
					    & 9943                      & 1439                      & 12.64 \\ \hline\hline

\multicolumn{7}{||c||}{\textbf{$\gamma$ = 0.005 and \textit{nu} = 0.15}}                                                                                                                                                                                                                                                                                                         \\ \hline
\multicolumn{1}{||c|}{} & \multicolumn{3}{c||}{Feasible Trajectories Set}                                                         & \multicolumn{3}{c||}{Non-Feasible Trajectories Set}                                                          \\ \hline
\multicolumn{1}{||c|}{} & \#correct                 & \# incorrect              & error (\%)                 & \#correct                 & \# incorrect               & error (\%)                 \\ \hline
\textit{poly}                    & 4005                      & 905                       & 18.10
					    & 9779                      & 1603                      & 14.08 \\ \hline
\textit{sigm}                 & 4262                      & 750                       & 14.96
					    & 9626                      & 1756                      & 15.43 \\ \hline\hline
\end{tabular}
\end{table}

The \textit{nu} parameter is a model configuration parameter and it refers to an upper bound on the fraction of training errors and a lower bound of the fraction of support vectors. Based on the results displayed in Table \ref{tab:1} one can verify that increasing the value of the \textit{nu} parameter decreases the error on the classification of unfeasible trajectories, while increasing the error on classifying feasible ones. Therefore, there is a trade-off on defining the best parametrization for the SVDD models.

It was found that the configuration with \(\gamma = 0.05\) and \textit{nu} = 0.15 produces the most balanced classification model, resulting in errors for classifying feasible and non-feasible trajectories of 14.98\% and 15.45\%, respectively. In the last iteration of the proposed approach, as depicted in Figure \ref{fig:flowchart}, the flexibility trajectories selected by the DSO or flexibility aggregator still have to be validated locally by the HEMS. In the case of a non-feasible flexibility trajectory selection, the HEMS is responsible of indicating the most similar feasible trajectory that it can provide as flexibility to the interested agent.

\section{Conclusions}
\label{conc}

This paper proposed a novel concept, called multi-period flexibility forecast for LV prosumers (behind-the-meter flexibility), which combines small-scale DER flexibility (e.g., storage, EWH) and forecast uncertainty for PV and net-load time series. This new function can be embedded in an HEMS and explored in the context of smart grid and microgrid technology. The flexibility potential also accounts for comfort constraint regarding the temperature of the water in the EWH tank and a costumer-defined mode of operation of the electric battery. 

In a first stage, an EPSO-based generation mechanism is proposed to create a set of feasible flexibility trajectories that are robust to net-load forecast uncertainty. Then, in a second phase, it explores a support vector data description function as a ``black-box'' model to communicate the flexibility set to different types of users, like DSO and flexibility aggregators. Nevertheless, the discrete representation with the flexibility trajectories set can be also integrated in management functions of DSO and market players.

The modified EPSO algorithm used for searching feasible flexibility trajectories showed a high diversity in producing feasible solutions and, consequently, improved the efficiency of the created SVDD models that are responsible of learning the HEMS feasible flexibility set boundaries. More than that, the forecast uncertainty regarding the HEMS net-load profile is fully considered by means of fitness evaluation of the computed solutions for various base net-load short-term scenarios which was one of the major gaps identified in the state of the art. Consequently, the proposed methodology can be used to control the robustness of the flexibility trajectories set in a probabilistic fashion (i.e., chance-constrained optimization).

It is crucial that the flexibility set, which can be transmitted to the interested stakeholders, clearly incorporates the physical and operating constraints of the DER and also the desired strategic modes of operation defined by the HEMS end-user. The trajectories resulting from the developed EPSO-based algorithm respect the problem constraints regarding DER power limits, state-of-charge of the battery and the temperature of the water inside the EWH tank. Additionally, the constraint related to the customer's preferences about the use of the battery during PV surplus periods is also respected by the generated flexibility trajectories.

Additionally, the performance of the SVDD model was assessed for the classification of feasible and non-feasible flexibility trajectories. Several configurations regarding the different hyper-parameters of the SVDD algorithm were analyzed. The configuration found to be more balanced regarding the trajectory classification procedure,  with \textit{sigmoid} kernel, $\gamma = 0.05$, and $\nu = 0.15$, resulted in a misclassification error of around 15\%. These errors were computed for a 16 time steps problem formulation. As the total number of time steps of the problem decreases, the errors also tend to decrease. The SVDD parameters showed a significant impact on the model classification performance, imposing a trade-off between misclassification of feasible and non-feasible trajectories.

The final outcome of this work is a forecast methodology with benefits for two groups of stakeholders: a) integrated in DSO operational planning tools to explore local flexibility potential within LV distribution grids in a time efficient fashion while complying with costumer privacy concerns; b)  used by a prosumers flexibility aggregator to efficiently assess the true flexibility provision capacity that can be provided by the assets belonging to its portfolio.

Future work will focus on improving the efficiency of the classification procedure by the SVDD model and also on investigating new approaches to transmit to the interested agents the HEMS flexibility set, e.g. use virtual batteries derived from the flexibility trajectories set.

\section*{Acknowledgements}
The research leading to this work is being carried out as a part of the InteGrid project (\textit{Demonstration of INTElligent grid technologies for renewables INTEgration and INTEractive consumer participation enabling INTEroperable market solutions and INTErconnected stakeholders}), which received funding from the European Union’s Horizon 2020 Framework Programme for Research and Innovation under grant agreement No. 731218.

The work of Rui Pinto was also supported in part by Funda\c{c}\~ao para a Ci\^encia e a Tecnologia (FCT) under PhD Grant SFRH/BD/117428/2016.

\appendix
\section{Gaussian Copula Method}
\label{app:gaussian}

The Gaussian copula method \cite{Cario1997,Pinson2009} generates $M$ temporal scenarios of solar power forecasts. First, the variable $Z_{t+k|t}  = \Phi ^{ - 1} \left( {\hat F_{t+k|t}\left( {y_{t+k} } \right)} \right)$ is calculated, where $y_{t+k}$ is the observed value for lead time $t+k$, $\hat F_{t+k}$ the forecasted probability distribution function and $\Phi ^{ - 1}$ the  inverse of the Gaussian cumulative distribution function. This random variable $Z_{t+k|t}$ is Gaussian distributed with zero mean and unit standard deviation. 
Then, the scenarios are generated with the following process:
\begin{enumerate}
\item Generate $M$ random vectors $\bf Z^{\left[ m \right]}$ from a multivariate Gaussian distribution (i.e., Gaussian copula) with zero mean and covariance matrix $\bf \Sigma _Z$. The size of this random vector is between $t+1$ and $t+T$ where $T$ is the maximum lead time or time horizon. 
\item Transform $\bf Z^{\left[ m \right]}$ with the following equation to obtain a random vector in the same scale of $y$ (note that $Z$ was a Gaussian variable).
\begin{equation} \label{eq:transf}
{\bf y}^{\left[ m \right]}  = {\bf \hat F}^{ - 1} \left( {\Phi \left( {{\bf Z}^{\left[ m \right]} } \right)} \right)
\end{equation}
where ${\bf \hat F}^{ - 1} $ is the vector of inverse forecasted distribution function for lead times between $t+1$ and $t+T$, $\Phi$ is the distribution function of a standard normal random variable and ${\bf y}^{\left[ m \right]}$ is the m-th solar power scenario.
\end{enumerate}
The dependency structure between the lead-times is modeled with a Gaussian copula that uses an exponential covariance matrix given by:
\begin{equation} \label{eq:exp_cov}
{\mathop{\rm cov}} \left( {Z_{t + k_1 } ,Z_{t + k_2 } } \right) = \exp \left( { - \frac{{\left| {k_1  - k_2 } \right|}}{\nu }} \right)
\end{equation}
where $Z_{t + k_1 }$ is the Gaussian random variable for lead time $t + k_{1}$ and where $\nu$ is the range parameter controlling the strength of the correlation of random variables among the set of lead times. The parameter $\nu$ is determined by trial-error experiences using the p-variogram score as a performance metric \cite{Scheuerer2015}.

\renewcommand\bibname{References}
\section*{References}
\bibliography{Elsevier_ref}

\end{document}